%% file: ms.tex
\documentclass[times,twocolumn,final,authoryear]{elsarticle_preprint}

\usepackage{ycviu_preprint}
\usepackage{framed,multirow}

\usepackage{amssymb}
\usepackage{latexsym}

\usepackage{amsmath}
\usepackage{url}
\usepackage{booktabs} 	
\usepackage{arydshln}	
\usepackage{xspace}
\usepackage[capbesideposition=outside,captionskip=5pt]{floatrow}		
\usepackage{pifont}
\usepackage{hyperref}
\usepackage{fancyhdr}  

\newcommand{\refsec}[1]{Sec.~\ref{sec:#1}}

\newcommand{\refeq}[1]{Eq.~\ref{eq:#1}}
\newcommand{\reffig}[1]{Fig.~\ref{fig:#1}}
\newcommand{\reftab}[1]{Tab.~\ref{tab:#1}}

\newcommand{\boldparagraph}[1]{\vspace{0em}\noindent{\bf #1}}

\newcommand*{\ie}{i.e.\@\xspace}
\newcommand{\cmark}{\ding{51}}%
\newcommand{\xmark}{\ding{55}}%
\definecolor{dec_blue}{RGB}{0, 90, 179}
\definecolor{enc_green}{RGB}{0, 110, 10}
\definecolor{lat_brown}{RGB}{140, 77, 30}
\def\cite#1{\citep{#1}}  

\journal{Computer Vision and Image Understanding}

\begin{document}

\clearpage

\ifpreprint
  \setcounter{page}{1}
\else
  \setcounter{page}{1}
\fi

\begin{frontmatter}

\title{Weakly Supervised Learning of Multi-Object 3D Scene Decompositions Using Deep Shape Priors}

\author[1,2,3]{Cathrin \snm{Elich}\corref{cor1}}
\cortext[cor1]{Corresponding author: 
  E-Mail: cathrin.elich@tuebingen.mpg.de\\}
\author[3]{Martin R. \snm{Oswald}}
\author[3,4]{Marc \snm{Pollefeys}}
\author[1]{Joerg \snm{Stueckler}}

\address[1]{Max Planck Institute for Intelligent Systems, Tuebingen, Germany}
\address[2]{Max Planck ETH Center for Learning Systems}
\address[3]{ETH Zurich, Zurich, Switzerland}
\address[4]{Microsoft Mixed Reality and AI Zurich Lab, Switzerland}

\received{27 October 2021}
\finalform{-}
\accepted{-}
\availableonline{-}
\communicated{-}

\begin{abstract}
	Representing scenes at the granularity of objects is a prerequisite for scene understanding and decision making.
	We propose PriSMONet, a novel approach based on \textbf{Pri}or \textbf{S}hape knowledge for learning \textbf{M}ulti-\textbf{O}bject 3D scene decomposition and representations from single images.
	Our approach learns to decompose images of synthetic scenes with multiple objects on a planar surface into its constituent scene objects and to infer their 3D properties from a single view.
	A recurrent encoder regresses a latent representation of 3D shape, pose and texture of each object from an input RGB image.
	By differentiable rendering, we train our model to decompose scenes from RGB-D images in a self-supervised way.
	The 3D shapes are represented continuously in function-space as signed distance functions which we pre-train from example shapes in a supervised way.
	These shape priors provide weak supervision signals to better condition the challenging overall learning task.
	We evaluate the accuracy of our model in inferring 3D scene layout, demonstrate its generative capabilities, assess its generalization to real images, and point out benefits of the learned representation.}  
\end{abstract}


\end{frontmatter}

\section{Introduction}

\begin{figure*}
	\includegraphics[width=.99\linewidth]{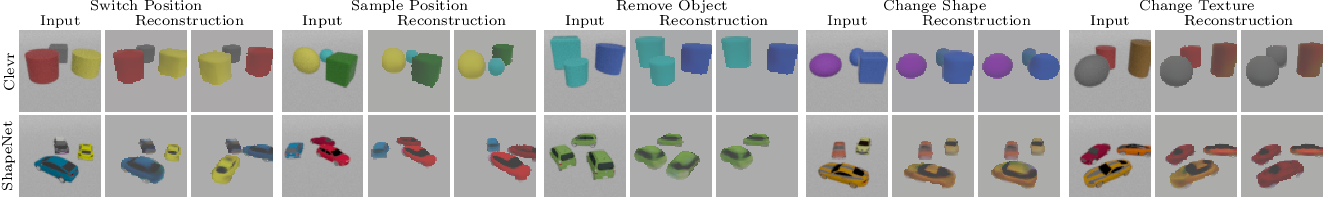}
	\caption{\textbf{Example scenes with object manipulation.}
	For each example, we input the left images to our network and obtain the reconstruction shown in the middle image.
	After the manipulation in the latent space, we obtain the respective right image.
	Plausible new scene configurations are shown on the Clevr dataset~\cite{johnson2017_clevr} (top) and on composed ShapeNet models~\cite{chang2015_shapenet} (bottom).}
	\label{fig:clevr_manipulate}
\end{figure*}

Humans have the remarkable capability to decompose scenes into their constituent objects and to infer object properties such as 3D shape and texture from just a single view.
Providing intelligent systems with similar capabilities is a long-standing goal in artificial intelligence. 
Such representations would facilitate object-level description, abstract reasoning and high-level decision making.
Moreover, object-level scene representations could improve generalization for learning in downstream tasks such as robust object recognition or action planning.
Learning single image 3D scene decomposition in a self-supervised way is specifically challenging due to common ambiguities with respect to depth, 3D object pose, shape, texture and lighting, for which suitable priors are required.

Previous work on learning-based scene representations focused on single-object scenes~\cite{sitzmann2019_srns}, did not consider the underlying compositional structure of scenes~\cite{eslami2018_gqn, mildenhall2020_nerf}, or neglected to model the 3D geometry of the scene and the objects explicitly with an interpretable representation (e.g.~\cite{burgess2019_monet,greff2019_iodine,eslami2016_air,stelzner2021_obsurf}).
In our work, we make steps towards multi-object representations by proposing a network which learns to decompose scenes into objects through weak and self-supervision, and represents 3D shape, texture, and pose of objects explicitly.
By this, our approach jointly addresses the tasks of object detection, instance segmentation, object pose estimation and inference of 3D shape and texture in single RGB images.
We incorporate prior shape knowledge in the form of pre-trained neural implicit shape models to allow for learning of scene decomposition into an interpretable 3D representation through weak supervision.

Inspired by~\cite{park2019_deepsdf,oechsle2019_texturefields}, we represent 3D object shape and texture continuously in function-space as signed distance and color values at continuous 3D locations.
The scene representation network infers the object poses and its shape and texture encodings from the input RGB image.
We use a differentiable renderer which efficiently generates color and depth images as well as instance masks from the object-wise scene representation.
This allows for training our scene representation network in a weakly-supervised way. 
Using a pre-trained shape space, we train our model to decompose and describe the scene without further annotations like instance segmentation, object poses, texture, or concrete shape from single RGB-D images.
Due to the combination of object-level scene understanding and differentiable rendering, our model further facilitates to generate new scenes by altering an interpretable latent representation (see Fig.~\ref{fig:clevr_manipulate}).

We evaluate our approach on both synthetic and real scene datasets with images composed of multiple objects on a planar background.
We show its capabilities with shapes such as geometric primitives and vehicles, and demonstrate the properties of our geometric and weakly-supervised learning approach for scene representation.

In summary, we make the following \textbf{contributions}: 
	\textbf{(1)} We propose PriSMONet, a novel model to learn representations of scenes composed of multiple objects with a planar background.
	Our model describes the scene by explicitly encoding object poses, 3D shapes and texture.
	\textbf{(2)} Our model is trained via differentiable rendering to decode the latent representation back into images.
	We apply a differentiable renderer using sampling-based raycasting for deep SDF shape embeddings which renders color and depth images as well as instance segmentation masks. 
	This setup enables our model to be trained using only weak supervision in form of shape priors and eliminates the need for scene specific object-wise 3D supervision.
	\textbf{(3)} By representing 3D geometry explicitly, our approach naturally respects occlusions between objects and facilitates manipulation of the scene within the latent space.
		 We demonstrate properties of our geometric model for scene representation and augmentation, and discuss advantages over multi-object scene representation methods which model 3D geometry implicitly.

To the best of our knowledge, our approach is the first to jointly learn object instance detection, instance segmentation, object localization, and inference of 3D shape and texture in a single RGB image via weak and self-supervised scene decomposition. 
For our current model, we make several assumptions and simplifications to provide insights for this challenging task and to allow for an in-depth evaluation of the applied strategies.
In particular, we train and test our model on synthetic scenes with uniformly colored, planar background, and simplified lighting conditions.
We also test our model trained with synthetic data on real images.
We provide a discussion about current limitations of our model and possible directions for future research in \refsec{limitations}.


\begin{figure*} 
	\centering
	\caption{\textbf{Multi-object 3D scene representation network.} The image is sequentially encoded into  \textcolor{lat_brown}{object representations} using an \textcolor{enc_green}{encoder network $g_0$}. The object encoders additionally receive image and mask compositions ($\Delta I, M$) generated from the previous object encodings. A differentiable renderer based \textcolor{dec_blue}{ decoder $F$} composes images and masks from the encodings of previous steps. The background is encoded from the image in parallel and used in the final scene reconstruction. }
	\label{fig:pipeline_full}
	\includegraphics[width=0.9\linewidth]{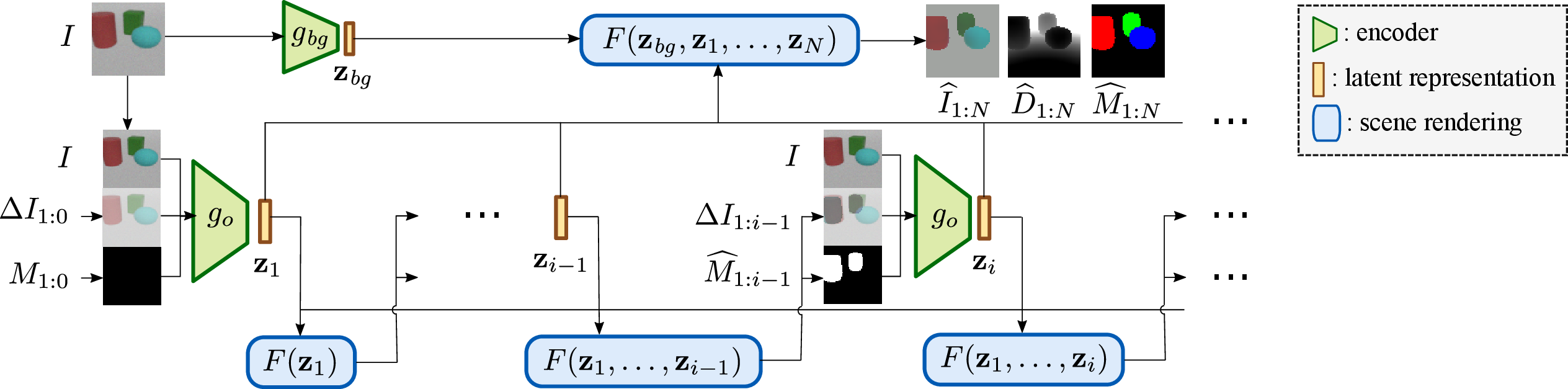}
\end{figure*}

\section{Related Work}


\boldparagraph{Deep learning of single object geometry.} 
Several recent 3D learning approaches represent single object geometry by implicit surfaces of occupancy or signed distance functions which are discretized in 3D voxel grids~\cite{kar2017_mvsmachine,tulsiani2017_diffraycons,wu2016_3dgan,gadelha2017_3dshapeind,qi2016_volumetric,rezende2016_unsup3dstructure,choy2016_3dr2n2,shin2018_pixels,Xie2019_Pix2Vox}.
Voxel representations typically waste significant memory and computation resources in empty scene parts. 
This limits their resolution and  capabilities to represent fine details.
Other methods represent shapes with point clouds~\cite{qi2017_pointnet,achlioptas2018_genpcmodel}, meshes \cite{groueix2018_atlasnet}, deformations of shape primitives~\cite{henderson2019_learning} or multiple views~\cite{tatarchenko2016_mv3dmodels}.
In continuous representations, neural networks are trained to directly predict signed distance~\cite{park2019_deepsdf,xu2019_disn,sitzmann2019_srns}, occupancy~\cite{mescheder2019_occnetworks,chen2019_imnet}, or texture~\cite{oechsle2019_texturefields} at continuous query points.
We use such representations for individual objects.


\boldparagraph{Deep learning of multi-object scene representations.}
Self-supervised learning of multi-object scene representations from images recently gained significant attention in the machine learning community. 
MONet~\cite{burgess2019_monet} presents a multi-object network which decomposes the scene using a recurrent attention network and an object-wise autoencoder. 
It embeds images into object-wise latent representations and overlays them into images with a neural decoder.
\cite{Yang2020_Learning} improve upon this work.
\cite{greff2019_iodine} use iterative variational inference to optimize object-wise latent representations using a recurrent neural network.
SPAIR~\cite{crawford2019_spair} and SPACE~\cite{lin2020_space} extend the attend-infer-repeat approach~\cite{eslami2016_air} by laying a grid over the image and estimating the presence, relative position, and latent representation of objects in each cell.
In GENESIS~\cite{engelcke2020_genesis}, the image is recurrently encoded into latent codes per object in a variational framework. 
\cite{locatello2020_slotattention} propose Slot Attention for decomposing scenes into objects.
In contrast to our method, the above methods do not represent the 3D geometry of the scene explicitly.

Related to our approach are also generative models like \cite{liao2020_3DContrImgSynthesis, nguyenphuoc2020_blockgan} which generate novel 3D scenes but do not explain input views like we do.
\mbox{GIRAFFE}~\cite{niemeyer2020_giraffe} proposes a generative model for scene composition based on neural radiance fields (NeRF~\cite{mildenhall2020_nerf}) which samples shape and appearance latents of objects.
Different to ours, the method does not decompose images into 3D object descriptions.
Recently,~\cite{stelzner2021_obsurf} decompose a scene into objects using Slot Attention and condition a NeRF-based decoder on a latent code to vary object shape and appearance.
Their model does encode object position and rotation implicitly and does not provide an explicit interpretable 3D parametrization like our method.
Other methods exploit multiple images to describe 3D scenes~\cite{henderson2020_3dVideoGen, nanbo2020_mulmon, chen2021_roots}.
Scene decomposition in 3D from a single view, however, is significantly more difficult and requires certain assumptions like prior shape knowledge to be trainable in a self-supervised way.


\boldparagraph{Supervised learning for object instance segmentation, pose and shape estimation.} Loosely related are supervised methods that segment object instances~\cite{ren2015_fasterrcnn, redmon2016_yolo, hou2019_3dsis}, estimate their poses~\cite{xiang2017posecnn} or recover their 3D shape~\cite{gkioxari2019_meshrcnn,Kniaz2020_Image}.
In Mesh R-CNN~\cite{gkioxari2019_meshrcnn}, objects are detected in bounding boxes and a 3D mesh is predicted for each object.
The method is trained supervised on images with annotated object shape ground truth.
In contrast to all of them, our method is trained without ground-truth annotations of object pose, segmentation masks, or appearance which our model learns with only weak supervision.


\boldparagraph{Neural and differentiable rendering.} 
\cite{eslami2018_gqn} encode images into latent representations which can be aggregated from multiple views.
Scene rendering is deferred to a neural network which is trained to decode the latents into images from examples.
Several differentiable rendering approaches have been proposed using voxel occupancy  grids~\cite{tulsiani2017_diffraycons,gadelha2017_3dshapeind,rezende2016_unsup3dstructure,yan2016_ptn,gwak2017_ws3d,zhu2018_von,wu2017_marrnet,nguyenphuoc2018_rendernet}, meshes~\cite{kato2018_neuralrenderer,loper2014_opendr,chen2019_dibrender,delaunoy2011_gradflows,ramamoorthi2001_invrendering,meka2018_lime,athalye2018_srae,richardson2016_facerecon,liu2019_softras,henderson2019_learning}, signed distance functions~\cite{sitzmann2019_srns}, or point clouds \cite{lin2018_learning,yifan2019_dss}.
\cite{henderson2016_2d3dmesh} apply differentiable rendering to learn textured 3D meshes of single objects from 2D images.
Recent literature overviews on differentiable rendering are \cite{tewari2020_neuralrendering,kato2020differentiable}.
In our work, we find depth and mask values through equidistant sampling 
along the ray.


\section{Method}

We propose an autoencoder architecture which embeds images into object-wise scene representations (see Fig.~\ref{fig:pipeline_full} for an overview).
Each object is explicitly described by its 3D pose and latent embeddings for both its shape and textural appearance.
Given the object-wise scene description, a decoder composes the images back from the latent representation through differentiable rendering.
We train our autoencoder-like network in a self-supervised way from RGB-D images.


\boldparagraph{Scene Encoding.}
The network infers a latent $\mathbf{z} = \left( \mathbf{z}_1, \ldots, \mathbf{z}_N, \mathbf{z}_{\mathit{bg}} \right)$ which decomposes the scene into object latents $\mathbf{z}_i \in \mathbb{R}^d$, $i \in \left\{ 1, \ldots, N \right\}$ and a background component $\mathbf{z}_{\mathit{bg}} \in \mathbb{R}^{d_{bg}}$ where $d, d_{bg}$ are the dimensionality of the object and background encodings and $N$ is the object count.
Objects are sequentially encoded by a deep neural network $\mathbf{z}_i = g_o( I, \Delta I_{1:i-1}, \widehat{M}_{1:i-1} )$ (see Fig.~\ref{fig:pipeline_full}).
We share the same object encoder network and weights between all objects.
To guide the encoder to regress the latent representation of one object after the other, we forward additional information about already reconstructed objects.
Specifically, we decode the previous object latents into object composition images, depth images and occlusion masks ${(\widehat{I}_{1:i-1}, \widehat{D}_{1:i-1}, \widehat{M}_{1:i-1}) := F( \mathbf{z}_{bg}, \mathbf{z}_1, \ldots, \mathbf{z}_{i-1} })$. 
They are generated by $F$ using differentiable rendering which we detail in the subsequent paragraph.
We concatenate the input image $I$ with the difference image ${\Delta I_{1:i-1} := I-\widehat{I}_{1:i-1}}$ and occlusion masks $\widehat{M}_{1:i-1}$, and input this to the encoder for inferring the representation of object~$i$.

\begin{figure}[t!]
	\centering
	\includegraphics[width=\linewidth]{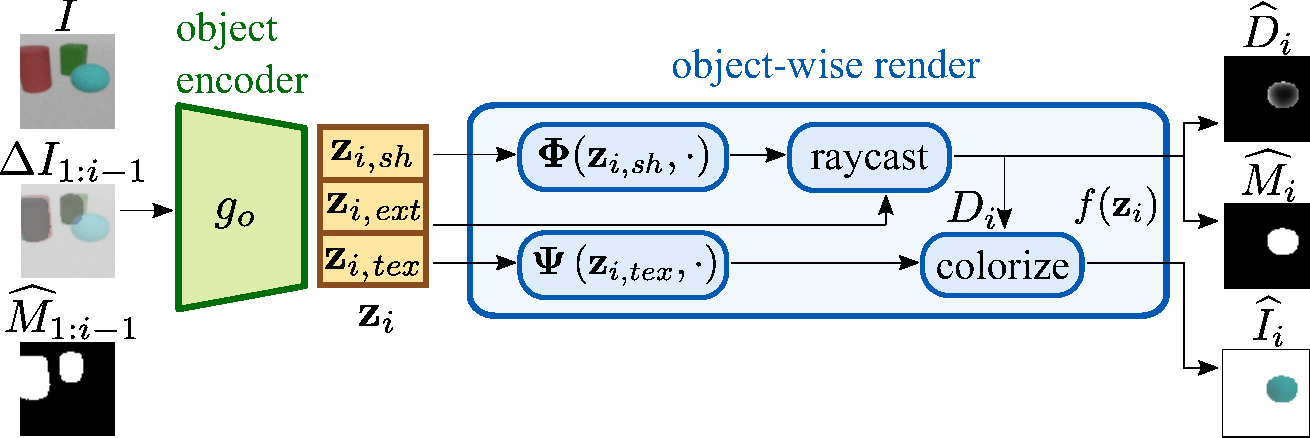}
	\caption{\textbf{Object-wise encoding and rendering.} We feed the input image, scene composition images and masks of the previously found objects to an \textcolor{enc_green}{object encoder network $g_o$} which regresses the \textcolor{lat_brown}{encoding of the next object~$\mathbf{z}_i$}. The object encoding decomposes into shape~$\mathbf{z}_{i,\mathit{sh}}$, extrinsics~$\mathbf{z}_{i,\mathit{ext}}$ and texture latents~$\mathbf{z}_{i,\mathit{tex}}$. The shape latent parametrizes an SDF function network~$\mathbf{\Phi}$ which we use in combination with the pose and scale of the object encoded in~$\mathbf{z}_{i,\mathit{ext}}$ for raycasting the object depth and mask using our \textcolor{dec_blue}{differentiable renderer $f$}. Finally, the color of the pixels is found with a texture function network~$\mathbf{\Psi}$ parametrized by the texture latent.}
	\label{fig:pipeline_object}
\end{figure}

The object encoding $\mathbf{z}_i = (\mathbf{z}_{i,\mathit{sh}}^\top, \mathbf{z}_{i,\mathit{tex}}^\top, \mathbf{z}_{i,\mathit{ext}}^\top)^\top$ decomposes into encodings for shape $\mathbf{z}_{i,\mathit{sh}}$, textural appearance $\mathbf{z}_{i,\mathit{tex}}$, and 3D extrinsics $\mathbf{z}_{i,\mathit{ext}}$ (see Fig.~\ref{fig:pipeline_object}).
The shape encoding $\mathbf{z}_{i,\mathit{sh}} \in \mathbb{R}^{D_{\mathit{sh}}}$ parametrizes the 3D shape represented by a DeepSDF autodecoder~\cite{park2019_deepsdf}.
Similarly, the texture is encoded in a latent vector $\mathbf{z}_{i,\mathit{tex}} \in \mathbb{R}^{D_{\mathit{tex}}}$ which is used by the decoder to obtain color values for each pixel that observes the object.
Object position $\mathbf{p}_i = (x_i,y_i,z_i)^\top$, orientation $\theta_i$ and scale $s_i$ are regressed with the extrinsics encoding $\mathbf{z}_{i,\mathit{ext}} = ( \mathbf{p}_i^\top, z_{\cos,i}, z_{\sin,i}, s_i )^\top$.
The object pose $\mathbf{T}_w^o( \mathbf{z}_{i,\mathit{ext}} ) =$~\scalebox{0.7}{$\left( \begin{array}{cc} s_i \mathbf{R}_i^\top & -\mathbf{R}_i^\top \mathbf{p}_i\\ \mathbf{0} & 1 \end{array} \right)$}
is parametrized in a world coordinate frame with known transformation~$\mathbf{T}_{c}^{w}$ from the camera frame. 
We assume the objects are placed upright and model rotations around the vertical axis with angle~$\theta_i = \arctan( z_{\sin,i}, z_{\cos,i} )$ and  corresponding rotation matrix $\mathbf{R}_i$.
We use a two-parameter representation for the angle as suggested in~\cite{zhou2019_controtrep}. 
We scale the object shape by the factor $s_i \in \left[ s_{\min}, s_{\max} \right]$ which we limit in an appropriate range using a sigmoid squashing function.
The background encoder $g_{\mathit{bg}} := \mathbf{z}_{\mathit{bg}} \in \mathbb{R}^{d_{bg}}$ regresses the uniform color of the background plane, i.e. $d_{bg}=3$.
We assume the plane extrinsics and hence its depth image is known in our experiments.


\boldparagraph{Scene Decoding.}
\label{sec:decoder}
Given our object-wise scene representation, we use differentiable rendering to generate individual images of objects based on their geometry and appearance and compose them into scene images. 
An object-wise renderer $(\widehat{I}_{i}, \widehat{D}_{i}, \widehat{M}_{i}) := f( \mathbf{z}_i )$ determines color image $\widehat{I}_i$, depth image $\widehat{D}_i$ and occlusion mask $\widehat{M}_i$ from each object encoding independently (see Fig.~\ref{fig:pipeline_object}).
The renderer determines the depth at each pixel $\mathbf{u} \in \mathbb{R}^2$ (in normalized image coordinates) through raycasting in the SDF shape representation.
Inspired by \cite{wang2020_directshape}, we trace the SDF zero-crossing along the ray by sampling points $\mathbf{x}_j := ( d_j\mathbf{u}, d_j)^\top$ in equal intervals $d_j := d_0 + j \Delta d, j \in \left\{ 0, \ldots, N-1 \right\}$ with start depth $d_0$.
The points are transformed to the object coordinate system  by
$\mathbf{T}_c^o( \mathbf{z}_{i,\mathit{ext}} ) := \mathbf{T}_w^o( \mathbf{z}_{i,\mathit{ext}} ) \mathbf{T}_c^w$.
Subsequently, the signed distance $\mathbf{\phi}_j$ to the shape at these transformed points is obtained by evaluating the SDF function network $\mathbf{\Phi}\left( \mathbf{z}_{i,\mathit{sh}}, \mathbf{T}_c^o( \mathbf{z}_{i,\mathit{ext}} ) \mathbf{x}_j \right)$. 
Note that the SDF network is also parametrized by the inferred shape latent of the object.
The algorithm finds the zero-crossing at the first pair of samples with a sign change of the SDF $\mathbf{\Phi}$.
The sub-discretization accurate location $\mathbf{x}(\mathbf{u})$ of the surface is found through linear interpolation of the depth regarding the corresponding SDF values of these points.
The depth at a pixel $D_i(\mathbf{u})$ is given by the z coordinate of the raycasted point $\mathbf{x}(\mathbf{u})$ on the object surface in camera coordinates.
If no zero crossing is found, the depth is set to a large constant.
The binary occlusion mask $M_i(\mathbf{u})$ is set to $1$ if a zero-crossing is found at the pixel and $0$ otherwise.
The pixel color $I_i(\mathbf{u})$ is determined using a decoder network $\mathbf{\Psi}$ similar to $\mathbf{\Phi}$ which receives the texture latent $\mathbf{z}_{i,\mathit{tex}}$ of the object and the raycasted 3D point $\mathbf{x}(\mathbf{u})$ in object coordinates as inputs and outputs an RGB value, \ie $I_i(\mathbf{u}) = \mathbf{\Psi}\left( \mathbf{z}_{i,\mathit{tex}}, \mathbf{T}_c^o( \mathbf{z}_{i,\mathit{ext}} ) \mathbf{x}(\mathbf{u}) \right)$ (cf. \cite{oechsle2019_texturefields}).
Note, that albeit object masks are binary and only specify at which pixels color and depth have been rendered for an object, the gradients flow through the rendered depth and colors.

We speed up the raycasting process by only considering pixels that lie within the projected 3D bounding box of the object shape representation.
This bounding box is known since the SDF function network is trained with meshes that are normalized to fit into a unit cube with a constant padding.
Note that this rendering procedure is implemented using differentiable operations making it fully differentiable for the shape, color and extrinsics encodings of the object.

The scene images, depth images and occlusion masks 
$\big(\widehat{I}_{1:n}, \widehat{D}_{1:n}, \widehat{M}_{1:n}\big) = F( \mathbf{z}_{\mathit{bg}}, \mathbf{z}_{1}, \ldots, \mathbf{z}_{n} )$
are composed from the individual objects $1,\ldots,n$ with $n\leq N$ and the decoded background through z-buffering.
We initialize them with the background color, depth image of the empty plane and empty mask.
Recall that the background color is regressed by the encoder network.
For each pixel $\mathbf{u}$, we search the occluding object $i$ with the smallest depth at the pixel.
If such an object exists, we set the pixel's values in $\widehat{I}_{1:N}, \widehat{D}_{1:N}, \widehat{M}_{1:N}$ to the corresponding values in the object images and masks.


\boldparagraph{Training.}
We use pre-trained deep SDF models as a shape prior in our approach which were trained from a collection of meshes from different object categories similar to~\cite{park2019_deepsdf}.
Note that the pre-trained shape space of multiple object categories is a very weak prior for object detection and object-wise scene decomposition which our model learns in a self-supervised manner.
Our multi-object network is trained from RGB-D images containing example scenes composed of multiple objects.
To this end, we minimize the total loss function

\begin{equation}
L_{\mathit{total}} = \lambda_{I} L_{I} + \lambda_{D} L_{D} + \lambda_{\mathit{gr}} L_{\mathit{gr}} + \lambda_{\mathit{sh}} L_{\mathit{sh}},
\end{equation}
which is a weighted sum of multiple sub-loss functions:

\begin{align*}
    L_{I} &= \frac{1}{\vert\Omega\vert}\sum_{\mathbf{u} \in \Omega} \left\| G\left(\widehat{I}_{1:N}\right)(\mathbf{u}) - G(I_{\mathit{gt}})(\mathbf{u}) \right\|^2 _2 
	\\
	L_{D} &= \frac{1}{\vert\Omega\vert}\sum_{\mathbf{u} \in \Omega} \left\| G\left(\widehat{D}_{1:N}\right) (\mathbf{u}) - G(D_{\mathit{gt}})(\mathbf{u}) \right\|_1
	\\
	L_{\mathit{gr}} &= \sum_i \max( 0, -z_{i} ) + \max ( 0, -\phi_{i}(z'_{i} ))
	\\
	L_{\mathit{sh}} &= \sum_i \| \mathbf{z}_{i,\mathit{sh}} \|^2
\end{align*}

\begin{figure*} 
	\includegraphics[width=.99\linewidth]{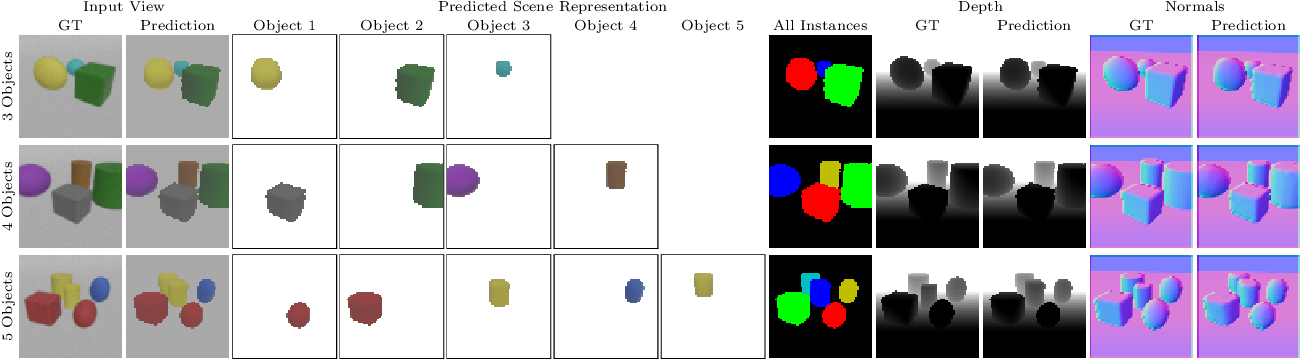}
	\caption{\textbf{Qualitative results on Clevr dataset~\cite{johnson2017_clevr}.} Our multi-object scene representation segments objects from the background and assigns object-wise instance label, geometry, appearance, and pose.}
	\label{fig:clevr_res}
\end{figure*}

In particular, $L_{I}$ is the mean squared error on the image reconstruction with $\Omega$ being the set of image pixels and $I_{\mathit{gt}}$ the ground-truth color image.
The depth reconstruction loss $L_{D}$ penalizes deviations from the ground-truth depth $D_{\mathit{gt}}$. 
We apply Gaussian smoothing $G(\cdot)$ to spread the gradients over the rendered image.
We decrease the standard deviation over time to allow the network to learn to decompose the scene in a coarse-to-fine manner.
$L_{\mathit{sh}}$ regularizes the shape encoding to stay within the training regime of the SDF network.
Lastly, $L_{\mathit{gr}}$ favors objects to reside above the ground plane with $z_{i}$ being the coordinate of the object in the world frame, $z'_{i}$ the corresponding projection onto the ground plane, and $\phi_{i}( \mathbf{x}_k ) := \mathbf{\Phi}\left( \mathbf{z}_{i,\mathit{sh}}, \mathbf{T}_c^o( \mathbf{z}_{i,\mathit{ext}} ) \mathbf{x}_k \right)$.
The shape regularization loss is scheduled with time-dependent weighting. This prevents the network from learning to generate unreasonable extrapolated shapes in the initial phases of the training, but lets the network refine them over time.

We use a CNN for both the object and the background encoder. 
Both consist of multiple convolutional layers with kernel size $(3, 3)$ and strides $(1, 1)$ each followed by ReLU activations and $(2, 2)$ max-pooling.
The subsequent fully-connected layers yield the encodings for objects and background.
Similar to \cite{park2019_deepsdf}, we use multi-layer fully-connected neural networks for the shape decoder $\mathbf{\Phi}$ and texture decoder $\mathbf{\Psi}$.
Further details are provided in the supplementary material.


\section{Experiments}


\boldparagraph{Datasets.} We provide extensive evaluation of our approach using synthetic scenes based on the Clevr dataset~\cite{johnson2017_clevr} and scenes generated with ShapeNet models~\cite{chang2015_shapenet}.
The Clevr-based scenes contain images with a varying number of colored shape primitives (spheres, cylinders, cubes) on a planar single-colored background.
We modify the data generation of Clevr in a number of aspects:
\textbf{(1)} We remove shadows and additional light sources and only use the Lambertian rubber material for the objects' surfaces as our decoder is by design not able to generate shadows.
\textbf{(2)} To increase shape variety, we apply random scaling along the principal axes of the primitives.
\textbf{(3)} An object might be completely hidden behind another one. Hence, the network needs to learn to hide superfluous objects.
We generate several multi-object datasets.
Each dataset contains scenes with a specific number of objects which we choose from two to five. 
Each dataset consists of $12.5\mathrm{K}$ images with a size of 64$\times$64 pixels. 
Objects are randomly rotated and placed in a range of $[-1.5, 1.5]^2$ on the ground plane while ensuring that any two objects do not intersect.
Additionally to the RGB images, we also generate depth maps for training as well as instance masks for evaluation.
The images are split into subsets of ($9\mathrm{K}/1\mathrm{K}/2.5\mathrm{K})$ examples for training, validation, and testing.
For the pre-training of the DeepSDF~\cite{park2019_deepsdf} network, we generate a small set of nine shapes per category with different scaling along the axes for which we generate ground truth SDF samples.
Different to~\cite{park2019_deepsdf}, we sample a higher ratio of points randomly in the unit cube instead of close to the surface.  
We also evaluate on scenes depicting either cars or armchairs as well as a  mixed set consisting of mugs, bottles and cans (tabletop) from the ShapeNet model set. 
Specifically, we select $25$ models per setting which we use both for pre-training the DeepSDF as well as for the generation of the multi-object datasets. 
We render ($18\mathrm{K}/2\mathrm{K}/5\mathrm{K}$) images per object category.
For additional evaluation, we further rendered an additional multi-object testset using $25$ previously unseen models.

\boldparagraph{Network Parameters.}
\label{sec:parameters}
For the Clevr / ShapeNet datasets, the object latent dimension is set to $D_{sh}=8 / 16$ and $D_{tex}=7 / 15$.
The shape decoder is pre-trained for $10\mathrm{K}$ epochs.
We linearly decrease the loss weight $\lambda_{sh}$ from $0.025 / 0.1$ to $0.0025 / 0.01$ during the first $500\mathrm{K}$ iterations.
The remaining weights are fixed to $\lambda_{I}=1.0$, $\lambda_{depth}=0.1 /0.05$, $\lambda_{gr}=0.01$.
We add Gaussian noise to the input RGB images and clip depth maps at a distance of $12$.
The renderer evaluates at 12 steps per ray.
Gaussian smoothing is applied with kernel size $16$ and linearly decreasing sigma from $\frac{16}{3}$ to $\frac{1}{2}$ in $250\mathrm{K}$ steps.
We trained models with ADAM optimizer~\cite{kingma2014_adam}, learning rate $10^{-4}$, and batch size 8 for $500/ 400$ epochs.
Training on the Clevr dataset with 3 objects takes about 2 days on a RTX2080Ti.


\input{tab4a_clevr_std.tex} 

\boldparagraph{Evaluation Metrics.}
We evaluate the learning of object-level 3D scene representations using measures for instance segmentation, image reconstruction, and pose estimation.
To evaluate our models' capability to recognize objects that best explain the input image, we consider established instance segmentation metrics.
An object is counted as correctly segmented if the intersection-over-union (IoU) score between ground truth and predicted mask is higher than a threshold $\tau$.
To account for occlusions, only objects that occupy at least $25$ pixels are taken into account.
We report average precision (AP\textsubscript{0.5}), average recall (AR\textsubscript{0.5}), and F1\textsubscript{0.5}-score for a fixed $\tau=0.5$ over all scenes as well as the mean AP over thresholds in range $[0.5, 0.95]$ with stepsize $0.05$ similar to \cite{Everingham2010_voc}.
We further list the ratio of scenes were all visible objects were found w.r.t. $\tau=0.5$ (allObj).

Next, we evaluate the quality of both the RGB and depth reconstruction of the generated objects.
To assess the image reconstruction, we report \textit{Root Mean Squared Error} (RMSE), \textit{Structural SIMilarity Index} (SSIM)~\cite{Wang2004_ImgQualAssessment}, and \textit{Peak Signal-to-Noise Ratio} (PSNR) scores~\cite{Wang2009_mse}.
For the object geometry, we compute similar to \cite{eigen2014_depthmappred} the \textit{Absolute Relative Difference} (AbsRD), \textit{Squared Relative Difference} (SqRD), as well as the RMSE for the predicted depth.
Furthermore, we report the error on the estimated objects' position (mean) and rotation (median, sym.: up to symmetries) for objects with a valid match w.r.t. $\tau=0.5$.
We show results over five runs per configuration and report the mean.


\subsection{Clevr Dataset}

In \reffig{clevr_res}, we show reconstructed images, depth and normal maps on the Clevr~\cite{johnson2017_clevr} scenes.
Our model provides a complete reconstruction of the individual objects although they might be partially hidden in the image.
The network can infer the color of the objects correctly and gets a basic idea about shading (e.g. that spheres are darker on the lower half).
The shape characteristics such as extent, edges or curved surfaces are well recognized.
As our model needs to fill all object slots, we sometimes observed that it fantasizes and hides additional objects behind others. 
Some reconstruction artifacts at object boundaries are due to rendering hard transitions between objects and background.


\begin{figure}[tb]
	\includegraphics[width=.8\linewidth]{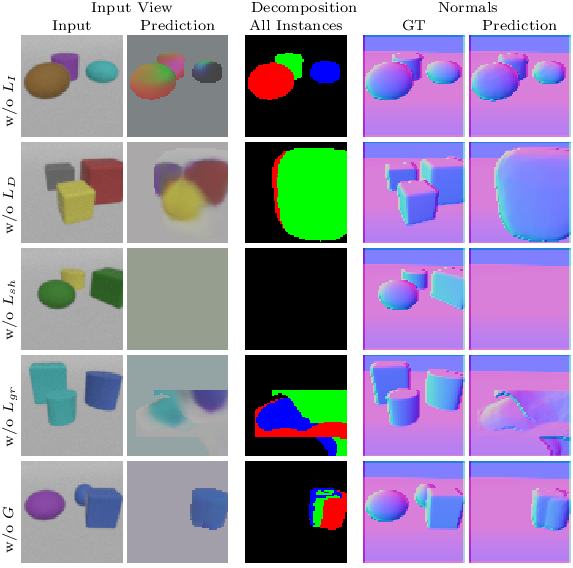}
	\caption{\textbf{Qualitative results for ablation study.} 
	Typical failure cases can be observed when leaving out individual components of our model.
	The combination of all or proposed loss functions is necessary to obtain a reasonable decomposition into the individual objects as well as meaningful object-wise representations which allow an appropriate scene reconstruction.}
	\label{fig:ablation}
\end{figure}

\begin{figure*} 
	\includegraphics[width=.99\linewidth]{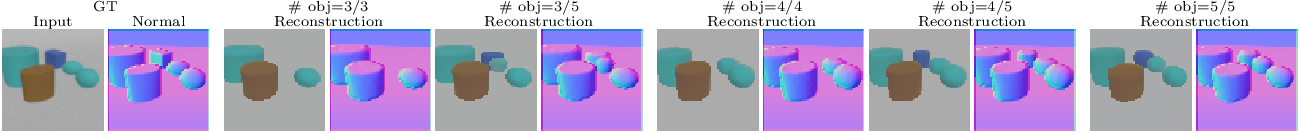}
	\caption{\textbf{Qualitative results on the Clevr dataset~\cite{johnson2017_clevr} with varied number of objects.} 
	As we use a shared encoder for detecting the objects in a recurrent architecture, it is possible to evaluate our model on a different number $o_{\mathit{test}}$ of objects than it was trained on ($o_{\mathit{train}}$). 
	For this, we reset the number of recurrent encoding steps to the number of objects in the test data. 
	We show reconstruction results for varying numbers $\#obj=o_{\mathit{train}}/o_{\mathit{test}}$.
	Remarkably, our models that were trained only on either three or four objects are able to recognize larger number of objects.}
	\label{fig:clevr_obj_count}
\end{figure*}

\boldparagraph{Ablation Study.} 
We evaluate various components of our model on the Clevr dataset with three objects.  
In Table~\ref{tab:clevr_ablation}, we compare training settings where we left out each of the loss functions. We further demonstrate the benefit of applying Gaussian smoothing (denoted by $G$), the importance of the additional input modalities as well as the effect of noise on depth maps.

The sequential encoder requires information about previously detected objects which are provided by the combined occlusion mask $\widehat{M}_{1:i-1}$ and difference image $\Delta I_{1:i-1}$. Without these, the model can only infer the same object prediction along all slots. While using only either of them provides enough information to guide the network in detecting missing objects, a combination of both works best in finding most objects (allObj).
At the beginning of training, the shape regularization loss is crucial to keep the shape encoder close to the pre-trained shape space and to prevent it from diverging due to the inaccurate pose estimates of the objects. 
Applying and decaying Gaussian blur distributes gradient information in the images beyond the object masks and allows the model to be trained in a coarse-to-fine manner.
This helps the model to localize the various objects in the scene.
The depth loss is essential for learning the scene decomposition. 
Without this loss, the network can simply describe several objects using a single object with more complex texture.
The usage of the ground loss prevents the model from fitting objects into the ground plane.
The image reconstruction loss plays only a minor part for the scene decomposition task but is merely responsible for learning the texture of the objects.
Visualizations of these findings can be seen in \reffig{ablation}.
Using all our proposed loss functions yields best results over all metrics.

We observe only a slight decrease in performance when training on noisy depth maps.
For this experiment, we added Gaussian noise with standard deviation $\sigma = \eta \cdot d^2$ to the depth maps ($\eta =0.001$, pixel-wise depth $d$).
This indicates, that our model is able to learn from non-perfect depth maps.
Remarkably, our model is able to find objects at high recall rates (0.942 AR at 50\% IoU).


\boldparagraph{Manipulation.} 
Our 3D scene model naturally facilitates generation and manipulation of scenes by altering the latent representation.
In~\reffig{clevr_manipulate}, we show example operations like switching the positions of two objects, changing their shape, or removing an entire object.
The explicit knowledge about 3D shape also allows us to reason about object penetrations when generating new scenes.
Specifically, we evaluate an object intersection loss $L_{int}$ on the newly sampled scenes to filter out those that turn out to be unrealistic due to an intersection between objects:
	\begin{equation}
	\label{eq:l_intersect}
		L_{\mathit{int}} = \sum_{i, j<i} \frac{1}{K} \sum_{k=1}^K \max(-(\phi_{i}( \mathbf{x}_k ) + \phi_{j}( \mathbf{x}_k )), 0) \enspace,
	\end{equation}
	where $i,j$ are object indices and $\mathbf{x}_k$ are $K$ sample points distributed evenly between the object centers.


\boldparagraph{Object Count.}
\label{sec:object_count}
We demonstrate generalization to different maximum numbers of objects in \reftab{clevr_ablation}. 
The model is trained with the respective number of objects in the dataset ($o_{train}$). 
Due to the setup of our dataset, it might happen that objects are occluded and thus not visible in the image. This enforces the model to learn to hide spare objects behind another one.
On average, our model finds and describes objects in less crowded scenes more easily, while it still performs with high accuracy for five objects.

Besides evaluating the trained networks on scenes with equal settings, we also examine its transferability to scenes with a different number of objects.
Due to the sequential architecture of our model, it can even be extended to parse scenes with more objects than it has been trained for ($o_{test}$).
As we use a shared encoder for all objects, we can simply reset the number of encoding rollouts to the number of objects in the test data.
Note that we assume the maximum number of objects to be known.
Although our model would be able to hide redundant objects behind already reconstructed ones without this explicit change, it cannot reconstruct additional objects.
Our model yields reasonable results, but performs best for similar object numbers in training and testing.
The achieved AR\textsubscript{0.5} and allObj measures indicate that the model is able to detect the objects at good rates.
For instance, for \#obj=3/5, our model finds 71\% of all objects (AR\textsubscript{0.5}) and can explain the full scene in about 21\% cases.
Qualitative results can be seen in \reffig{clevr_obj_count}.

\begin{figure}[tb]
	\includegraphics[width=.8\linewidth]{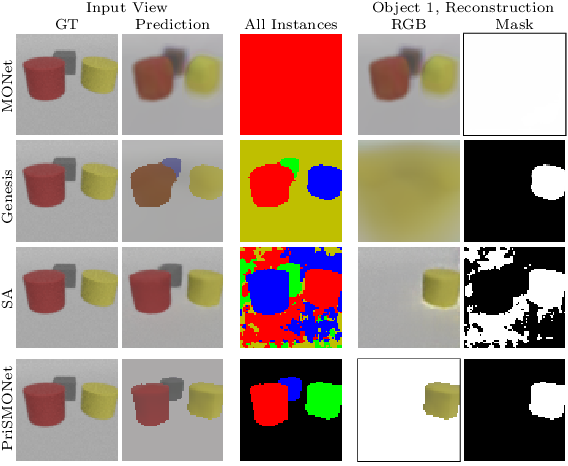}
	\caption{\textbf{Comparison to 2D Baselines}.
	The used implementation of MONet showed difficulties to decompose the scenes and, instead, the network would describes an entire scene using a single slot only.
	Genesis is able to decompose the scene into objects and separates them cleanly from the background. However, reconstruction results are weaker than ours.
	While Slot Attention (SA) yields a good RGB reconstruction, it often mixes object masks with the background.  
	Due to explicit rendering of 3D shapes, our model naturally differentiates between individual objects and background.
    }
	\label{fig:baseline}
\end{figure}

\boldparagraph{Comparison to 2D Baselines.}
We compare our method to the 2D multi-object scene representation approaches MONet \cite{burgess2019_monet}, Genesis \cite{engelcke2020_genesis}, and Slot Attention (SA) \cite{locatello2020_slotattention} in \reftab{baseline} and \reffig{baseline}.
We used provided code\footnote{Genesis, MONet: \url{https://github.com/applied-ai-lab/genesis}, SA: \url{https://github.com/google-research}} (adapted  for 64x64 images and \#objs+bg slots) with original hyperparameters for the original Clevr setup and trained it on our dataset.
In case of SA, we obtained masks by assigning each pixel to the slot with highest decoded alpha value.
For evaluation, we use both our metrics and the \textit{Adjusted Rand Index} (ARI) \cite{rand71_ari, hubert85_ari} which measures clustering similarity and was used in \cite{locatello2020_slotattention}. 
We consider both the full ARI score and their variant limited to the ground truth foreground pixels (ARI-FG).

\input{tab4b_clevr_baselines.tex}

Our experiments with MONet did not yield any decomposition as the model would simply use a single object slot to describe the entire scene.
SA's low instance segmentation scores result from a high number of background pixels in the object masks which becomes especially clear when comparing the high difference in performance for ARI and ARI-FG.
Genesis is able to decompose the scene into objects but reconstruction are worse than SA or ours.
Due to the usage of shape priors, our model is naturally restricted to produce a reasonable foreground/ background decomposition.
In contrast to our method, none of the others estimate any 3D information (e.g. shape or pose).
Furthermore, their object representation is not interpretable and does not allow intuitive manipulation of the scene.


\begin{figure*}[tb] 
	\includegraphics[width=.99\linewidth]{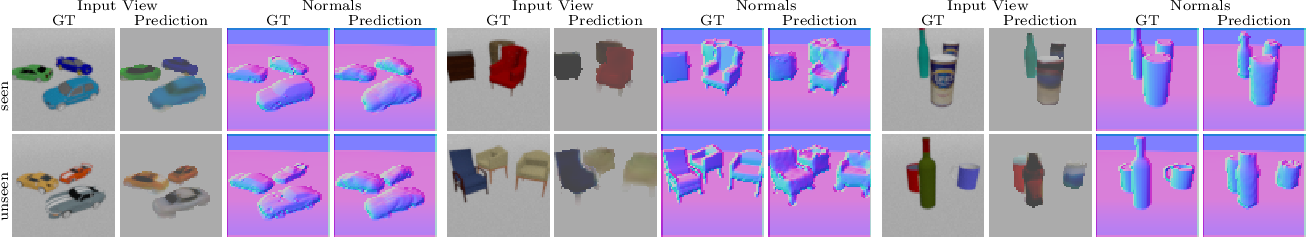}
	\caption{\textbf{Qualitative results on ShapeNet~\cite{chang2015_shapenet}.} Our model obtains a good scene understanding also with more difficult objects (cars, armchairs), handles different categories (tabletop scenes with mugs, bottles and cans), and estimates plausible poses, shapes and textures.}
	\label{fig:shapenet_res}
\end{figure*}

\input{tab4c_shapenet_std.tex}

\subsection{ShapeNet Dataset}

Our composed multi-object variant of ShapeNet~\cite{chang2015_shapenet} models is more difficult in shape and texture variation than Clevr~\cite{johnson2017_clevr}.
For some object categories such as cups or armchairs, training can converge to local minima.
We report mean and best results over five training runs in \reftab{shapenet}, where the best run is chosen according to F1 score on the validation set.
Evaluation is performed on two different test sets: scenes containing (1) object instances with shapes and textures used for training and (2) unseen object instances.
We show several scene reconstructions in \reffig{shapenet_res}.

For the cars, our model yields consistent performance in all runs with comparable decomposition results to our Clevr experiments.
However, we found that cars exhibit a pseudo\nobreakdash-180\nobreakdash-degree shape symmetry which was difficult for our model to differentiate.
Especially for small objects in the background, it favors to adapt the texture over rotating the object.
For the armchair shapes, our model finds local minima in pseudo\nobreakdash-90\nobreakdash-degree symmetries.
The median rotation error indicates better than chance prediction for the correct orientation.
Rotation error histograms can be found in the supplementary material.
For approximately correct rotation predictions, we found that our model was able to differentiate between basic shape types but often neglected finer details like thin armrests which are difficult to differentiate in the images.

Our tabletop dataset provides another type of challenge: the network needs to distinguish different object categories with larger shape and scale variation. 
For this setting, we added further auxiliary losses to penalize object intersections (\refeq{l_intersect}) as well as object positions outside of the image view:
\begin{equation}
	L_{p}= \sum_i \max( -\min( x^p_{i}, w-x^p_{i} ), 0 )
\end{equation}
Our model is able to predict the different shape types with coarse textures.
On scenes with instances that were not seen during training, our model often approximates the shapes with similar training instances.
As can be expected, results are slightly worse compared to the evaluation on shapes known from training.
Nevertheless, our model is still able to generate a reasonable scene decomposition using similar objects from the training set which demonstrates the generalization capability of our network.

\begin{figure}[tb]
	\includegraphics[width=.99\linewidth]{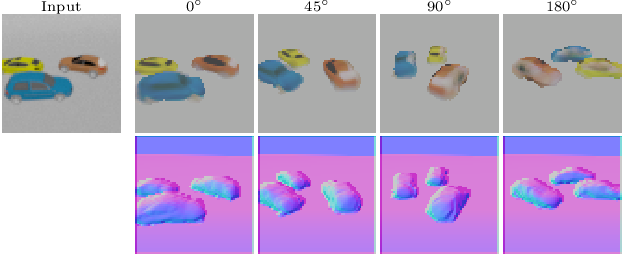}
	\caption{\textbf{Novel view renderings.} Our model is able to generate new scene renderings for largely rotated camera views from just a single input RGB image.
	While we noticed a reduced texture accuracy for unseen object parts, the normal maps demonstrate that our model obtains a good 3D structural understanding of the scene.
	\label{fig:shapenet_newview}}
\end{figure}

\boldparagraph{Novel Views.}
Due to the learned 3D structure, our model is able to render novel views from a scene given a single image (see \reffig{shapenet_newview}).
Although our model never saw multiple views of the same scene during training and is not tuned for this task, we obtain reasonable results for both scene geometry and appearance. 
We observe a lower texture reconstruction quality for invisible scene parts.

\boldparagraph{Supervised Training.}
We examine the benefits of using additional supervision for training.
Specifically, we utilize ground truth annotations for either \textbf{(1)} 3D object poses or \textbf{(2)} 2D foreground/ background segmentation masks (\reftab{supervised}).

For the first variant, we consider known 3D position, rotation around z-axis, and scale.
To account for object order invariance, we determine \textit{object matches} $(z_{i, ext}, z^{gt}_{m(i), ext})$ where each predicted object is assigned to a ground truth object such that every ground truth object is matched exactly once and the summed Euclidean distance between pairwise predicted and ground truth object is minimal.
With $\mathbf{z}_{i,\mathit{ext}} = ( \mathbf{p}_i^\top, \theta_i, s_i )^\top$, we use the following additional loss function during training:
\begin{equation}
	L_{pose} = \sum_{i} l_{pos}(\mathbf{p}_i, \mathbf{p}^{gt}_{m(i)}) + l_{rot}(\theta_i, \theta^{gt}_{m(i)}) + l_{scale}(s_i, s^{gt}_{m(i)}),
\end{equation}
where
${l_{pos}(\mathbf{p}_i, \mathbf{p}_j) =  || \mathbf{p}_i-\mathbf{p}_j ||_2}$, 
${l_{rot}(\theta_i, \theta_j) = 1-\text{cos}(\theta_i-\theta_j)}$, and  
${l_{scale}(s_i, s_j) = (s_i-s_j)^2.}$
We observe that supervision on ground truth 3D object poses helps our model over all categories to reliably decompose the scene into the constituent objects and to achieve improved accuracy on the pose estimation. 
We also note that this type of supervision helps our model to overcome local minima due to pseudo-symmetry.
The main drawback of using 3D poses for supervision is that this kind of annotation for real 2D images is very expensive.

For the second variant, we consider the combined foreground masks $\widehat{M}_{1:N}$, $M_{gt}$ for predicted and ground truth objects, apply Gaussian smoothing like for the image and depth reconstruction losses, and use binary cross entropy for computing the loss:
\begin{align}
	L_{mask} = \frac{1}{|\Omega|}\sum_{\mathbf{u} \in \Omega} 
	&G(M_{gt})(\mathbf{u})\log (G(\widehat{M}_{1:N})(\mathbf{u})) + \\
	&(1-G(M_{gt}))(\mathbf{u})\log (1-G(\widehat{M}_{1:N})(\mathbf{u}))\nonumber
\end{align}
This loss significantly helped for the tabletop dataset and also yielded improvements for car objects regarding the RGB and depth reconstruction measures compared to the unsupervised setup.
In contrast, the performance on the chair dataset decreased.
Especially, we observed that our model often only was able to detect two of the three objects and missed smaller objects in the background which leads to a low  AR\textsubscript{0.5} score.
This indicates that supervision on the foreground mask does not yield a sufficient training signal to always overcome local minima. 
However, this kind of supervision can still be interesting due to lower cost for annotation.

\input{tab4d_shapenet_supervised.tex}


\input{tab4e_shapenet_6dof.tex}

\boldparagraph{Extension to full 6 degree of freedom (DoF) position and rotation.}
While our main dataset considers a physically plausible setup where objects are placed stable on the ground, we further evaluate the reliance of our model on these assumptions and how it deals with an extended scene setup.
For this, we generate additional datasets with either Clevr or car objects where we lower the ground plane and allow objects to be placed at a height within $[-1.5, 1.5]$ as well as to have arbitrary 3D rotation.

We train our model on variants of this new datasets where either one or both of the previous assumptions are removed.
To enable the model to learn full rotation, we extend the extrinsic encoding with an axis-angle representation, $\mathbf{z}_{\mathit{ext}} = ( \mathbf{p}^\top, z_{\cos}, z_{\sin}, \mathbf{z}_{rot}, s )^\top$, 
where $\mathbf{z}_{rot} = (z_{rot, x}, z_{rot, y}, z_{rot, z})$ is a unit vector describing the axis of rotation. 
Please note that our base model is in principle already able to place objects at arbitrary heights as it predicts the 3D center of the objects.

Results can be seen in \reftab{flying}.
We observe that it is easier for our model to decompose scenes with fully rotated objects (at one height) compared to those where objects are placed at arbitrary heights (but rotated around the vertical axis only).
However, recognizing the full orientation is still more difficult compared to the original setup which results in weaker reconstruction results.
We further notice a higher chance to miss objects if these are placed at a variation of height positions which leads to a stronger performance decrease over the entire task.
While our model achieves decent results on this more difficult setup, future work on more difficult scenes is required.
Rotation errors are difficult to assess and hence omitted due to the various pseudo-symmetries for cars and actual symmetries in Clevr-object for full 3D rotation.
We provide qualitative results in the supplementary material.


\subsection{Real Data}
\label{sec:real}
\begin{figure}
	\includegraphics[width=.99\linewidth]{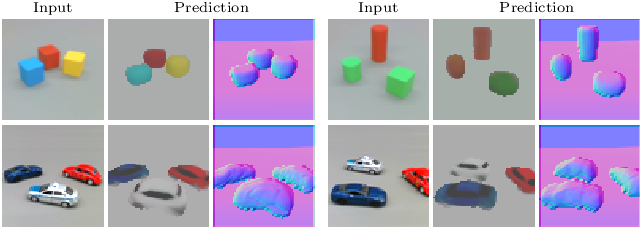}
	\caption{\textbf{Evaluation on real images.} We show results on real images by our model that was trained on synthetic data. We notice that our model is able to capture the coarse scene layout and shape properties of the objects. However, challenges arise due to domain, lighting, camera intrinsics and view point changes indicating interesting directions for future research.}
	\label{fig:real_images}
\end{figure}

\begin{figure}[tb]
	\includegraphics[width=.99\linewidth]{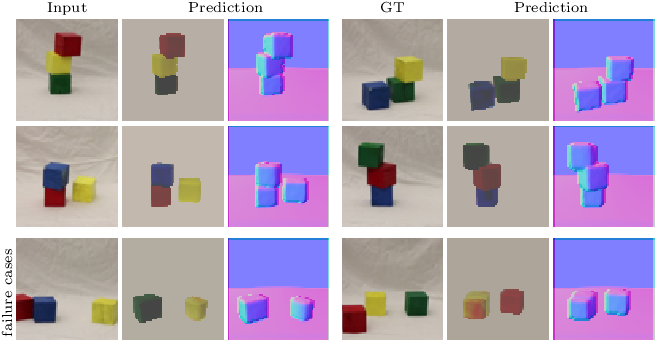}
	\caption{\textbf{Parsing real images of block towers  \cite{Lerer2016_BlockTowers}}. 
	We trained our model on synthetic images of stacked cubes and test on real images.
	Our model recognizes the scene configuration well, but
	occasionally objects are missed, especially if they are close to the image boundary.
	}
	\label{fig:real_images_blocktowers}
\end{figure}

We further evaluated our model on real images of toy cars and wooden building blocks (see \reffig{real_images})  as well as on the real block tower dataset from \cite{Lerer2016_BlockTowers} (see \reffig{real_images_blocktowers}). 
For the former dataset, we adjusted brightness and contrast of the photos to visually match the background color of the synthetic data.
For the block tower dataset, images were cropped and scaled.
Despite different camera and image properties, our model decomposes the scenes into objects and obtains their coarse shape and appearance without any domain adaptation or fine-tuning on real data.
Typical observed failure cases include wrong color prediction, difficulties with elongated shapes, and sometimes unrealistic object clusters.
Difficulties in reconstructing the objects correctly can be explained by the limited variety in the training data (e.g. there is no 'light green' texture in the Clevr dataset). Applying domain adaption or domain randomization might be interesting directions for future research.

\subsection{Limitations}
\label{sec:limitations}
\begin{figure}[tb]
	\includegraphics[width=.99\linewidth]{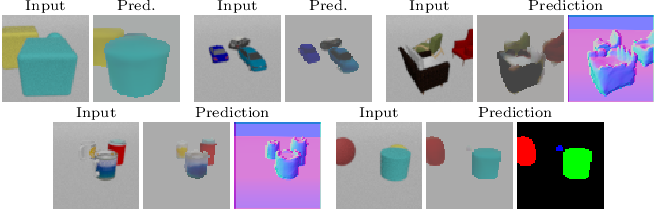}
	\caption{\textbf{Limitations}. Input and output pairs for typical failure cases and limitations of our method due to ambiguities for self-supervised learning. See text for details.}
	\label{fig:limitations}
\end{figure}

We show typical failure cases of our approach in Fig.~\ref{fig:limitations}.
Self-supervised learning without regularizing assumptions leads typically to ill-conditioned problems.
We use a pre-trained 3D shape space to confine the possible shapes, impose a multi-object decomposition of the scene, and use a differentiable renderer of the latent representation.
In our self-supervised approach, ambiguities can arise due to the decoupling of shape and texture.
For instance, the network can choose to occlude the background partially with the shape but fix the image reconstruction by predicting background color in these areas.
Rotations can only be learned up to a pseudo-symmetry by self-supervision when object shapes are rotationally similar and the subtle differences in shape or texture are difficult to differentiate in the image.
In such cases, the network can favor to adapt texture over rotating the shape.
Depending on the complexity of the scenes and the complex combination of loss terms, training can run into local minima in which objects are moved outside the image or fit the ground plane.
Currently, the network is trained for a maximum number of objects.
If all objects in the scene are explained, it hides further objects which could be alleviated by learning a stop criterion.
While the network is able to interpolate between shapes of the prior shape space seen during training, it cannot extrapolate to unknown shapes, for example, from unseen object categories.

\section{Conclusion}

We propose a novel deep learning approach for self-supervised multi-object scene representation learning and parsing.
Our approach infers the 3D structure of a scene from a single RGB image by recursively parsing the image for shape, texture and poses of the objects.
A differentiable renderer allows images to be generated from the latent scene representation and the network to be trained self-supervised from RGB-D images.
We employ pre-trained shape spaces that are represented by deep neural networks using a continuous function representation as an appropriate prior for this ill-posed problem.

Our experiments demonstrate the ability of our model to parse scenes with various object counts and shapes.
We provide an ablation study to motivate design choices and discuss assumptions and limitations of our approach.
We show the advantages of our model to reason about the underlying 3D space of a seen scene by performing explicit manipulation on the individual objects or rendering novel views.
While using synthetic data allows us to evaluate the design choices of our model in a controlled setup, we also show successful reconstructions of real images.
We believe our approach provides an important step towards self-supervised learning of object-level 3D scene parsing and generative modeling of complex scenes from real images.
Our work is currently limited to scenes with few objects as well as simple backgrounds and lighting conditions.
Future work will address the challenges of more complex scenes.

\section*{Acknowledgments}
This work has been supported by Cyber Valley, the Max Planck Society and Innosuisse funding (Grant No. 34475.1 IP-ICT).
We are grateful to the Max Planck ETH Center for Learning Systems for supporting Cathrin Elich.
We thank Michael Strecke for his support with generating our ShapeNet dataset.

\bibliographystyle{model2-names}
\bibliography{ms}

%

\end{document}

%% file: tab4a_clevr_std.tex
\begin{table*}
	\scriptsize
	\centering
	\setlength{\tabcolsep}{1pt}
	\newcommand{\gcs}{\hspace{7pt}} 
	\caption{\textbf{Results on the Clevr dataset~\cite{johnson2017_clevr}.} The combination of our proposed loss with Gaussian blur is essential to guide the learning of scene decomposition and object-wise representations. 
	We highlight best (bold) results for each measure among the full model and the variations where we left out individual components for ablation.
	Specifying the maximum numbers of objects, we further train our model on scenes with 2, 4, or 5 objects.
	Despite the increased difficulty for a larger number, our model recognizes most objects in scenes with two to five objects.
	Models trained with fewer objects can successfully explain scenes with a larger number of objects (\textit{\# obj=o\textsubscript{train}/o\textsubscript{test}}).}
	\label{tab:clevr_ablation}
	\begin{tabular}{lc@{\gcs}cccccc@{\gcs}cccc@{\gcs}cccc@{\gcs}c}
		\toprule
		&& \multicolumn{5}{c}{Instance Reconstruction} && \multicolumn{3}{c}{Image Reconstruction} && \multicolumn{3}{c}{Depth Reconstruction} && \multicolumn{1}{c}{Pose Est.} \\
		\cmidrule(l){2-7} \cmidrule(l){8-11} \cmidrule(l){12-15} \cmidrule(l){16-17}
	 	&& mAP $\uparrow$ & AP\textsubscript{0.5} $\uparrow$ & AR\textsubscript{0.5} $\uparrow$ & F1\textsubscript{0.5} $\uparrow$ & allObj $\uparrow$  && RMSE $\downarrow$ & PSNR $\uparrow$ & SSIM $\uparrow$ && RMSE $\downarrow$ & AbsRD $\downarrow$ & SqRD $\downarrow$ && Err\textsubscript{pos} \\
		\midrule
	 	\# obj=3/3, input: $(I)$ && 0.716 & 0.931 & 0.326 & 0.481 & 0.005 && 0.100 & 20.177 & 0.818 && 1.142 & 0.075  &0.292 && 0.150 \\
	 	\# obj=3/3, input: $(I, \Delta I_{1:i-1})$ && 0.715 & 0.951 & 0.878 & 0.903 & 0.712 && 0.054 & 25.716 & 0.904 && 0.585 & 0.022 & 0.070 && 0.154 \\
	 	\# obj=3/3, input: $(I, \widehat{M}_{1:i-1})$ && \textbf{0.719} & \textbf{0.953} & 0.927 & 0.935 & 0.817 && 0.050 & 26.375 & \textbf{0.914} && \textbf{0.554} & 0.020 & \textbf{0.061} && \textbf{0.151} \\
		\# obj=3/3, w/o $L_I$ &&  0.686 &  0.941 &  0.879 &  0.899  & 0.709 && 0.199 & 14.176 & 0.713 &&  0.595 &  0.023 &  0.073 && 0.159\\
	 	\# obj=3/3, w/o $L_D$ && 0.023 & 0.086 & 0.076 & 0.078 & 0.008 &&  0.085 &  22.142 &  0.837 && 2.745 & 0.231 & 1.061 && 1.341\\
	 	\# obj=3/3, w/o $L_{sh}$ && 0.01 & 0.032 & 0.027 & 0.028 & 0.001 && 0.13 & 17.907 & 0.763 && 1.455 & 0.147 & 0.556 && 0.676 \\
	 	\# obj=3/3, w/o $L_{gr}$ && 0.09 & 0.195 & 0.205 & 0.198 & 0.008 && 0.09 & 21.163 & 0.799 && 1.159 & 0.087 & 0.32 && 0.81 \\
	 	\# obj=3/3, w/o $G$ && 0.164 & 0.296 & 0.161 & 0.199 & 0.001 && 0.114 &  19.065 & 0.792 && 1.331 & 0.112 & 0.441 && 0.182 \\[0.3pt] \hdashline \noalign{\vskip 2pt}
	 	\# obj=3/3, noisy depth && 0.703 & 0.945 & 0.910 & 0.922 & 0.771 && 0.052 & 25.978 & 0.907 && 0.575 & 0.025 & 0.066 && 0.157\\[0.3pt] \hdashline \noalign{\vskip 2pt}
	 	\# obj=3/3, full [PriSMONet] && 0.712 & 0.949 & \textbf{0.942} & \textbf{0.943} & \textbf{0.850} && \textbf{0.049} & \textbf{26.466} & \textbf{0.914} && \textbf{0.554} & \textbf{0.019} & \textbf{0.061} && 0.155\\
	 	\midrule
	 	\# obj=2/2 && 0.782 & 0.977 & 0.963 & 0.967 & 0.928 && 0.039 & 28.389 &  0.941 && 0.432 & 0.012 & 0.04 && 0.138 \\
	 	\# obj=4/4 && 0.688 & 0.941 & 0.919 & 0.926 & 0.746 && 0.054 & 25.632 &  0.899 && 0.584 & 0.022 & 0.064 && 0.151 \\
	 	\# obj=5/5 && 0.604 & 0.895 & 0.861 & 0.872 & 0.539 && 0.061 & 24.568 &  0.876 && 0.593 & 0.025 & 0.067 && 0.149 \\[0.5pt] \hdashline \noalign{\vskip 1pt}
	 	\# obj=3/2 && 0.756 & 0.974 & 0.969 & 0.97 & 0.942 && 0.041 & 28.011 & 0.937 && 0.452 & 0.013 & 0.044 && 0.14\\
	 	\# obj=3/4 && 0.613 & 0.883 & 0.853 & 0.863 & 0.512 && 0.06 & 24.669 & 0.88 && 0.665 & 0.028 &  0.083 && 0.179\\
	 	\# obj=3/5 && 0.478 & 0.775 & 0.71 & 0.735 & 0.212 && 0.072 & 23.093 & 0.841 && 0.69 & 0.033 & 0.086 && 0.201\\	 	
	 	\bottomrule
	\end{tabular}
\end{table*}

%% file: tab4b_clevr_baselines.tex
\begin{table}[tb]
	\scriptsize 
	\centering
	\setlength{\tabcolsep}{1pt}
	\newcommand{\gcs}{\hspace{7pt}}  
		{\begin{tabular}{lc@{\gcs}lc@{\gcs}ccccc@{\gcs}cc@{\gcs}c}
		\toprule
		&&&& \multicolumn{4}{c}{Instance Rec.} && \multicolumn{1}{c}{RGB Rec.} && \multicolumn{1}{c}{3D Pred.} \\
		\cmidrule(l){4-8} \cmidrule(l){9-10} \cmidrule(l){11-12} 
	 	&&&& AP\textsubscript{0.5} $\!\uparrow$ & AR\textsubscript{0.5} $\!\uparrow$ & ARI $\!\uparrow$ & ARI-FG $\!\uparrow$  && RMSE $\!\downarrow$ \\
		\midrule 	
		\multirow{4}{*}{\rotatebox{90}{ Clevr}} 
		&& MONet && 0.000 & 0.000 & 0.000 & 0.002 && 0.058 && \xmark\\
		&& Genesis && 0.880 & 0.848 & 0.812 & 0.717 && 0.064 && \xmark\\
		&& Slot Attention && 0.089 & 0.099 & 0.088 & \bf 0.951 && \bf 0.017 && \xmark\\
		&& PriSMONet && \bf 0.949 & \bf 0.942 & \bf 0.891 & 0.798 && 0.049 && \cmark \\
	 	\bottomrule
		\end{tabular}}
	\caption{\textbf{Comparison to 2D Baselines}.
	Genesis shows decent results on both the decomposition and reconstruction task but is overall weaker than our method.
	SA performs better on RGB reconstruction, but worse on most instance segmentation measures because many background pixels are assigned to object slots while our model naturally differentiates objects and background.
	The used implementation MONet failed in decomposing the scene into the individual objects.
	In contrast to ours, none of the baseline methods do predict any explicit 3D information.
    }
	\label{tab:baseline}
\end{table}

%% file: tab4c_shapenet_std.tex
\begin{table*}[tb]
	\scriptsize
	\centering
	\setlength{\tabcolsep}{1pt}
	\newcommand{\gcs}{\hspace{7pt}}  
	\caption{\textbf{Evaluation on scenes with ShapeNet objects~\cite{chang2015_shapenet}.} Results for scenes containing objects from different categories. We differentiate between scenes that consist of shapes that were seen during training and novel objects. We show mean and best outcome over five runs.}
	\label{tab:shapenet}
	\begin{tabular}{lllc@{\gcs}cccccc@{\gcs}cccc@{\gcs}cccc@{\gcs}cl}
	\toprule
	&& && \multicolumn{5}{c}{Instance Reconstruction} && \multicolumn{3}{c}{Image Reconstruction} && \multicolumn{3}{c}{Depth Reconstruction}&& \multicolumn{2}{c}{Pose Estimation} \\
	\cmidrule(l){4-9} \cmidrule(l){10-13} \cmidrule(l){14-17} \cmidrule(l){18-20}
	&& && mAP\newline $\uparrow$ & AP\textsubscript{0.5} $\uparrow$ & AR\textsubscript{0.5} $\uparrow$ & F1\textsubscript{0.5} $\uparrow$ & allObj $\uparrow$ && RMSE $\downarrow$ & PSNR $\uparrow$ & SSIM $\uparrow$ && RMSE $\downarrow$ & AbsRD $\downarrow$ & SqRD  $\downarrow$ && Err\textsubscript{pos} $\downarrow$ & Err\textsubscript{rot} [sym.] $\downarrow$ \\
	\midrule
		\multirow{4}{*}{\rotatebox{90}{cars}} & \textit{seen} & best && 0.750 & 0.991 & 0.991 & 0.991 & 0.979 && 0.064 & 24.092 & 0.898 && 0.158 & 0.006 & 0.004 && 0.144 & 23.67$^{\circ}$ [3.29$^{\circ}$] \\
&                    & mean  && 0.738 & 0.990 & 0.990 & 0.990 & 0.975 && 0.064 & 23.979 & 0.894 && 0.160 & 0.006 & 0.005 && 0.146 & 22.09$^{\circ}$ [3.07$^{\circ}$] \\
	&   \textit{unseen}~ & best && 0.639 & 0.980 & 0.980 & 0.980 & 0.955 && 0.077 & 22.442 & 0.843 && 0.210 & 0.010 & 0.008 && 0.183 & 24.24$^{\circ}$ [4.53$^{\circ}$]\\
		&                & mean   && 0.632 & 0.977 & 0.977 & 0.977 & 0.944 && 0.077 & 22.454 & 0.842 && 0.208 & 0.010 & 0.008 && 0.184 &  24.25$^{\circ}$ [4.41$^{\circ}$] \\
		[0.5pt]\hdashline \noalign{\vskip 1pt}
		\multirow{4}{*}{\rotatebox{90}{chairs}} & \textit{seen} & best && 0.432 & 0.897 & 0.871 & 0.881 & 0.640 && 0.086 & 21.576 & 0.803 && 0.829 & 0.040 & 0.117 && 0.308 & 43.64$^{\circ}$ [9.13$^{\circ}$]\\
&                    & mean  && 0.329 & 0.642 & 0.638 & 0.640 & 0.188 && 0.102 & 20.137 & 0.772 && 1.021 & 0.058 & 0.196 && 0.296 & 55.12$^{\circ}$  [7.25$^{\circ}$]\\
	&   \textit{unseen}~ & best && 0.377 & 0.852 & 0.821 & 0.833 & 0.534 && 0.092 & 20.994 & 0.778 && 0.890 & 0.052 & 0.137 && 0.395 &  58.79$^{\circ}$ [10.66$^{\circ}$]\\
		&                & mean   && 0.278 & 0.613 & 0.607 & 0.609 & 0.158 && 0.106 & 19.740 & 0.746 && 1.068 & 0.069 & 0.213 && 0.372 & 68.29$^{\circ}$ [9.28$^{\circ}$] \\
		[0.5pt] \hdashline \noalign{\vskip 1pt}
		\multirow{4}{*}{\rotatebox{90}{tabletop}} & \textit{seen} & best && 0.628 & 0.936 & 0.870 & 0.895 & 0.659 && 0.057 & 25.242 & 0.908 && 0.786 & 0.026 & 0.132 && 0.182 & 89.14$^{\circ}$  \\
&                    & mean  && 0.394 & 0.565 & 0.537 & 0.546 & 0.251 && 0.078 & 22.871 & 0.861 && 1.022 & 0.050 & 0.231 && 0.155 & 88.53$^{\circ}$ \\

	&   \textit{unseen}~ & best && 0.435 & 0.839 & 0.816 & 0.823 & 0.569 && 0.083 & 21.807 & 0.840 && 1.034 & 0.044 & 0.224 && 0.275 & 89.25$^{\circ}$ \\
		&                & mean   && 0.285 & 0.530 & 0.521 & 0.523 & 0.237 && 0.102 & 20.160 & 0.800	 && 1.172 & 0.061 & 0.291 && 0.238 & 89.99$^{\circ}$ \\
	\bottomrule	
	\end{tabular}
\end{table*}

%% file: tab4d_shapenet_supervised.tex
\begin{table}[tb]
	\scriptsize 
	\centering
	\setlength{\tabcolsep}{1pt}
	\newcommand{\gcs}{\hspace{2pt}}  
	\begin{tabular}{llc@{\gcs}cccc@{\gcs}ccc@{\gcs}cc@{\gcs}cc}
		\toprule
		&&&  \multicolumn{3}{c}{Instance Rec.} && \multicolumn{2}{c}{RGB Rec.} && \multicolumn{1}{c}{Depth Rec.} && \multicolumn{2}{c}{Pose Est.} \\
		\cmidrule(l){3-6} \cmidrule(l){7-9}  \cmidrule(l){10-11}  \cmidrule(l){12-14} 
	 	&&& mAP $\!\uparrow$ & AP\textsubscript{0.5} $\!\uparrow$ & AR\textsubscript{0.5} $\!\uparrow$ && RMSE $\!\downarrow$ & PSNR $\!\uparrow$ && RMSE $\!\downarrow$ && Err\textsubscript{pos} $\downarrow$ & Err\textsubscript{rot}  \\
		\midrule 	
		\multirow{3}{*}{\rotatebox{90}{cars}} 
	 	& PriSMONet && 0.738 & \textbf{0.990} & \textbf{0.990}  && \textbf{0.064} & 23.979 && 0.160  && 0.146 & 22.09$^{\circ}$  \\		
		& + 3D pose && 0.745 & 0.988 & 0.988 && 0.068 & 23.567 && 0.160 && \textbf{0.071} & \textbf{7.28$^{\circ}$} \\
	 	& + 2D mask && \textbf{0.756} & \textbf{0.990} & \textbf{0.990} && \textbf{0.064} & \textbf{24.030} && \textbf{0.152}  && 0.133 & 21.00$^{\circ}$ \\
	 	[1pt]\hdashline \noalign{\vskip 2pt}
	 	\multirow{3}{*}{\rotatebox{90}{chairs}} 
	 	& PriSMONet && 0.329 & 0.642 & 0.638 && 0.102 & 20.137 && 1.021 && 0.296 & 55.12$^{\circ}$ \\
	 	& + 3D pose && \textbf{0.533} & \textbf{0.928} & \textbf{0.928} && \textbf{0.085} & \textbf{21.709} && \textbf{0.753} &&  \textbf{0.126} & \textbf{10.06$^{\circ}$} \\
	 	& + 2D mask && 0.302 & 0.559 & 0.561 && 0.106 & 19.788 && 1.088 && 0.290 & 35.31$^{\circ}$ \\
	 	[1pt]\hdashline \noalign{\vskip 2pt}
	 	\multirow{3}{*}{\rotatebox{90}{tabletop}} 
	 	& PriSMONet && 0.394 & 0.565 & 0.537 && 0.078 & 22.871 && 1.022 && 0.155 & 88.53$^{\circ}$ \\ 	
	 	& + 3D pose && 0.667 & \textbf{0.956} & \textbf{0.944} && \textbf{0.054} & 25.679 && 0.652 &&  \textbf{0.099}& 54.81$^{\circ}$ \\
	 	& + 2D mask && \textbf{0.676} & 0.953 & 0.942 && \textbf{0.054} & \textbf{25.780} && \textbf{0.638}  &&  \textbf{0.099} & \textbf{46.53$^{\circ}$} \\
	 	\bottomrule
	\end{tabular}
	\caption{\textbf{Supervised Training}.
		We compare our weakly supervised model to variants were we used either 3d object poses or 2D masks for additional supervision.
		Overall, additional supervision on 3D poses provides a stable training setup where nearly all objects are recognized.
		The usage of 2D foreground masks only partly improved results.
		}
	\label{tab:supervised}
\end{table}

%% file: tab4e_shapenet_6dof.tex
\begin{table}
	\scriptsize 
	\centering
	\setlength{\tabcolsep}{1pt}
	\newcommand{\gcs}{\hspace{2pt}}  
	\begin{tabular}{lc@{\gcs}cccc@{\gcs}ccc@{\gcs}cc@{\gcs}c}
		\toprule
		&&  \multicolumn{3}{c}{Instance Rec.} && \multicolumn{2}{c}{RGB Rec.} && \multicolumn{1}{c}{Depth Rec.} && \multicolumn{1}{c}{Pose Est.} \\
		\cmidrule(l){2-5} \cmidrule(l){6-8}  \cmidrule(l){9-10}  \cmidrule(l){11-12} 
	 	&& mAP $\!\uparrow$ & AP\textsubscript{0.5} $\!\uparrow$ & AR\textsubscript{0.5} $\!\uparrow$ && RMSE $\!\downarrow$ & PSNR $\!\uparrow$ && RMSE $\!\downarrow$ && Err\textsubscript{pos} $\downarrow$ \\
		\midrule 	
		 Clevr (standard data) && 0.712 & 0.949 & 0.942 && 0.049 & 26.466 && 0.553 \textit{(1.521)} && 0.155 \\	
		\enspace + rnd. height && 0.510 & 0.825 & 0.785 && 0.075 & 22.825 && 1.610 \textit{(3.448)} && 0.511\\
		\enspace + full 3D rot (*) && 0.610 & 0.925 & 0.922 && 0.065 & 23.995 && 1.420 \textit{(3.795)} && 0.264\\	
		\enspace + height, 3D rot (*) && 0.471 & 0.829 & 0.808 && 0.077 & 22.544 && 1.622 \textit{(3.414)} && 0.567\\	 
		[1pt]\hdashline \noalign{\vskip 2pt}
		Cars (standard data) && 0.738 & 0.990 & 0.990 && 0.064 & 23.979 && 0.160 \textit{(0.462)} && 0.146  \\	
		\enspace + rnd. height && 0.478 & 0.858 & 0.778 && 0.090 & 21.167 && 1.567 \textit{(2.893)} && 0.287\\	
		\enspace + full 3D rot (*) && 0.405 & 0.810 & 0.797 && 0.095 & 20.706 && 1.662 \textit{(3.101)} && 0.370 \\	
		\enspace + height, 3D rot (*) && 0.241 & 0.629 & 0.572 && 0.108 & 19.546 && 1.887 \textit{(2.802)} && 0.764\\	
	 	\bottomrule
	\end{tabular}
	\caption{\textbf{Extended 6DoF object poses.}
	We train our model on different dataset variants with less assumptions about the object poses.
	While it is more complicated compared to our main datasets to predict arbitrary 3D position and rotation, our approach is still able to decompose the scene with good accuracy according to AP\textsubscript{0.5} and AR\textsubscript{0.5} in most variants. Position and depth estimation degrade when object height and rotation are less constraint.
	(*) indicates an extension of our model which predicts objects' orientations with an axis-angle representation.
	The changed background in the adapted dataset impacts the evaluation of the depth reconstruction.
	For reference, we thus further list the error resulting from evaluating on the empty background $(\cdot)$.  
	Note that this error would even increase for objects placed at wrong positions.
	}
	\label{tab:flying}
\end{table}